\documentclass[journal]{IEEEtran}

\usepackage{cite}
\usepackage{url}
\usepackage{booktabs}
\usepackage{amsfonts}
\usepackage{amsmath,amssymb}
\usepackage{csvsimple}
\usepackage{enumitem}
\usepackage{tikz}
\usetikzlibrary{positioning}

\begin{document}

\title{%
Inhibitor Transformers and Gated RNNs
for Torus Efficient Fully Homomorphic Encryption
}

\author{Rickard~Br{\"a}nnvall,
    Tony~Zhang,
    Henrik~Forsgren,
    Andrei~Stoian,
    Fredrik~Sandin,
    Marcus~Liwicki%
\thanks{R. Br{\"a}nnvall, T. Zhang, and H. Forsgren were with RISE, Research Institutes of Sweden, Lule{\aa}, Sweden; A. Stoian was with Zama, Machine Learning Group, Paris, France; R. Br{\"a}nnvall, F. Sandin, and M. Liwicki were with Lule{\aa} University of Technology, Lule{\aa}, Sweden.}%
}

\markboth{Technical Report}{}

\maketitle

\begin{abstract}
This paper introduces efficient modifications to neural network-based sequence processing approaches, laying new grounds for scalable privacy-preserving machine learning under Fully Homomorphic Encryption (FHE).
Transformers are now ubiquitous in AI applications and have largely supplanted Gated Recurrent Neural Networks (RNNs) as the standard architecture for sequence modeling.
Both architectures rely on costly multiplications and complex activations that hinder encrypted inference.
We focus on TFHE, which supports deep circuit evaluation and efficient univariate function evaluation but makes variable-to-variable multiplication particularly expensive. To address this, we propose inhibitor designs for Transformers and gated RNNs that replace multiplications and Softmax/Sigmoid activations with additive and ReLU-based operations.
These changes enable integer-only computation, reduce circuit depth, and improve the efficiency of encrypted execution while preserving learning capacity.
We present complexity analyses and scaling experiments that indicate significant reductions in circuit depth and execution time under TFHE, with 3--6 times speedup for encrypted inference and 30--50\% reductions in plaintext inference time. 
Empirical evaluations on MNIST, IMDB, and IAM handwriting show inhibitor-based models maintain competitive accuracy. Knowledge distillation further demonstrates that an inhibitor-based DistilBERT achieves performance close to that of the conventional attention model on GLUE, positioning these architectures as a viable approach for scalable, privacy-preserving AI.
\end{abstract}


\section{Introduction}
\IEEEPARstart{T}{he} Transformer architecture ~\cite{Vasvani2017} has become one of the most influential designs in modern AI, achieving widespread adoption across natural language processing ~\cite{BERT,gpt2019,gpt2020}, computer vision ~\cite{ViT2020,swin2021,carion2020,perceiver2021}, and multimodal tasks~\cite{flamingo2022,stablediffusion2022,imagen2022}. 
Generative models such as Stable Diffusion~\cite{stablediffusion2022} and Imagen~\cite{imagen2022} leverage Transformer backbones for high-quality image synthesis. 
At the frontier, foundation models such as LLaMA~\cite{llama2023}, Mistral~\cite{mistral2023}, Gemini~\cite{gemini2023}, and Claude~\cite{claude2023} exemplify the trend toward hundred-billion-parameter scales, motivating research into smaller feature sizes and efficient architectures~\cite{Han2015,Han2016,Hinton2015}, reduced-precision arithmetic and quantization~\cite{Zhou2016,Jacob_2018,Krishnamoorthi2018}, for resource-constrained deployment~\cite{Horowitz_2014}.

The ability of Transformers to model long-range dependencies and exploit parallel computation has largely displaced earlier sequence models. Gated recurrent neural networks (RNNs), including LSTM~\cite{Hochreiter_1997} and GRU~\cite{Chung_2014}, were once the standard for temporal reasoning. 
Simple RNNs had suffered from exploding and vanishing gradients~\cite{Mikolov_thesis_2012,Pascanu_2013,Hochreiter_1991,Bengio_1994,Hochreiter_2009}, prompting the introduction of gating mechanisms that enabled long-range memory and achieved success in machine translation~\cite{Bahdanau2015} and handwriting recognition~\cite{GravesLiwicki2009}.
Recent extensions, such as xLSTM~\cite{Beck_xLSTM_2024}, introduce exponential gating and new memory structures to improve scalability, while hybrid approaches, such as HRM~\cite{HRM2024}, combine Transformer contextual richness with recurrence for multi-level reasoning.

\paragraph*{Challenge}
Both Transformers and gated RNNs rely heavily on variable-to-variable multiplications and on complex activation functions such as softmax and sigmoid. These operations are computationally expensive under Fully Homomorphic Encryption (FHE)~\cite{Gentry_2010}, a cryptographic technology that enables privacy-preserving machine learning by allowing computation directly on encrypted data. 
In this work, we focus on TFHE (Fully Homomorphic Encryption on the Torus) \cite{Chillotti_2019}, which enables deep circuit evaluation and efficient function computation, making it suitable for complex neural networks but costly for multiplications and non-linear activations.
This motivates the need for architectural modifications that reduce such costly operations without compromising the learning capacity on important tasks.

\paragraph*{Approach}
We propose an inhibitor mechanism for Transformers and gated RNNs. It replaces dot-product attention with the Manhattan distance in Transformers and substitutes multiplicative gating with subtractive gating in RNNs. The softmax and sigmoid activation functions are replaced with ReLU. These modifications are co-designed to preserve the expressive capacity of the original architectures while enabling TFHE-friendly integer-only computation, reducing circuit depth, and improving efficiency in encrypted environments.

This work makes four contributions
\begin{description}
    \item[Inhibitor Design:] We introduce the design of inhibitor attention for Transformers and inhibitor-gated RNNs, providing a principled alternative to conventional mechanisms.
    \item[Learning Capacity:] We demonstrate that these architectures maintain competitive learning capacity compared to standard models across diverse benchmarks.
    \item[Complexity Analysis:] We present an analysis showing significant reductions in circuit depth under TFHE, highlighting the suitability of our approach for privacy-preserving computation.
    \item[Empirical Validation:] We validate the proposed designs through extensive experiments in both plaintext and encrypted inference settings, confirming substantial efficiency gains without compromising accuracy.
\end{description}

\paragraph*{Paper Organization}
The remainder of this paper is organized as follows. Section~\ref{sec:background} reviews background and related work on privacy-preserving machine learning, Fully Homomorphic Encryption (FHE), and computational constraints relevant to encrypted inference. Section~\ref{sec:method} introduces the proposed inhibitor mechanisms for Transformers and gated RNNs, detailing their design principles and computational complexity. Section~\ref{sec:experiments} describes the experimental setup and machine learning benchmark tasks used to evaluate learning capacity, including knowledge distillation for compressed Transformer models. Section~\ref{sec:analysis} presents scaling analyses, comparing computational complexity and execution time under plaintext and encrypted execution for conventional and inhibitor-based designs under varying precision and sequence lengths. Section~\ref{sec:discussion} discusses key findings, mechanistic insights, limitations, and practical implications for privacy-preserving and resource-constrained environments. Finally, Section~\ref{sec:conclusion} concludes with a summary of contributions and directions for future research.

Preliminary versions of this work were presented at conferences and workshops: early plaintext and encrypted experiments for the Inhibitor Gated RNN appeared in~\cite{poster2023gatedRNN_unpublihed,poster2024_PPAI_gatedRNN}; for the Inhibitor Transformer in~\cite{poster2023inhibitor, poster2024TransformerFHE}; and for Knowledge Distillation in~\cite{poster2025inhibidistilbert}. This paper consolidates and extends those contributions with comprehensive evaluations and scaling analyses.

\section{Background and Related Work}
\label{sec:background}

This section summarizes key concepts and prior work, including Fully Homomorphic Encryption on the Torus (TFHE), privacy-preserving machine learning, and computational efficiency considerations for encrypted neural network inference.

\subsection{Computational Efficiency.}
Different operations within neural networks have distinct power and energy requirements, which can vary depending on the underlying computer architecture \cite{Parhami_book_2010}. 
Additions are inexpensive and typically executed in a single instruction, whereas multiplications are more costly. Operations involving literal constants are less expensive than variable-to-variable multiplications because constants can be encoded in program instructions, while variables require additional memory access, significantly impacting energy and execution time~\cite{Horowitz_2014}.

Activation functions play an essential role in neural networks. Functions such as sigmoid and softmax require exponential function evaluations and divisions, which are computationally expensive on conventional digital hardware. In contrast, the ReLU (Rectified Linear Unit) activation function has simpler implementations involving threshold comparisons, enabling more efficient execution.

Reduced-precision arithmetic and quantization~\cite{Zhou2016, Jacob_2018, Krishnamoorthi2018} project weights, activations, and inputs from floating-point to low-bit integers, supporting integer-only computation. These techniques reduce memory footprint and bandwidth while accelerating execution through efficient fixed-point operations. 

\subsection{Fully Homomorphic Encryption on the Torus}

TFHE~\cite{Chillotti_2019} introduces programmable bootstrapping (PBS), a mechanism that refreshes ciphertext to reduce noise while enabling function evaluation. PBS performs a blind rotation over a lookup table populated with values of a function on a discrete set of points, allowing efficient evaluation of non-linear operations. Compared to bootstrapping in other FHE schemes, PBS is relatively fast.

TFHE does not natively have multiplication between ciphertexts, but it can be constructed by the application of PBS operations
\begin{equation}
    a\, b = \mathrm{PBS}\left(f; a+b\right)
         - \mathrm{PBS}\left(f; a-b\right)
    \label{eq:mult_by_PBS_construction}
\end{equation}
where by $\mathrm{PBS}\left(f; x\right)$, we mean applying a table that corresponds to the function $f$ with argument $x$, which is 
\begin{equation}
    f(x) = \frac{x^2}{4}
    \label{eq:mult_by_PBS_function}
\end{equation}
for the case of multiplication. As it requires two PBS, multiplying two encrypted variables under TFHE can be thousands of times more expensive than an addition or literal multiplication, and therefore dominates the computational cost.

Division can be constructed by first applying PBS with $g(x)=1/x$ to obtain the reciprocal of the divisor, followed by multiplication using the PBS; practical implementations require quantization and safeguards for division by zero.

\subsection{Privacy Preserving Machine Learning.}
Variants of Fully Homomorphic Encryption (FHE) have enabled privacy-preserving machine learning (PPML) across classical and deep learning models. Early approaches targeted simple classifiers and nearest-neighbor methods~\cite{Bost2015MachineLC, Chakraborty_2022}, while initial neural network solutions such as Cryptonets~\cite{CryptoNets2016} used leveled HE with polynomial activations but faced scalability limitations.

Recent advances leverage TFHE \cite{Chillotti2016} for fast bootstrapping and look-up table evaluation. FHE–DiNN \cite{Bourse2018_FHEofDDNN} demonstrated discretized neural networks with linear complexity and accurate encrypted inference. SHE \cite{SHE2019} further optimized convolutional networks using ReLU, max pooling, and logarithmic quantization, achieving competitive accuracy on MNIST and CIFAR-10 with reduced latency.

TFHE-based approaches have extended PPML capabilities by improving efficiency and scalability. Architectures optimized for TFHE constraints employ programmable bootstrapping to enable encrypted inference in deep neural networks \cite{Stoian2023}. Quantized neural networks have been applied to medical image analysis under TFHE, achieving high accuracy with low latency \cite{Selvakumar2025}. 
TFHE has also been used for tree-based models, providing near-cleartext accuracy and practical inference times \cite{Frery2024}. 
To address performance bottlenecks, GPU-accelerated TFHE bootstrapping achieves up to 20 times speedup, enabling near real-time encrypted inference \cite{Xiao2025}. Nevertheless, inference on encrypted data remains orders of magnitude slower than plaintext computation, rendering large state-of-the-art Transformer models impractical.

\section{Methodology}
\label{sec:method}

Building on the challenges outlined in Section~\ref{sec:background}, this section introduces inhibitor-based modifications to Transformers and gated RNNs. These designs address the high cost of multiplications and complex activations under TFHE by replacing them with addition and ReLU operations.

\subsection{Transformer Neural Networks}

\begin{figure}
\centering
\resizebox{\columnwidth}{!}{%
    \begin{tikzpicture}[>=latex, thick]

\tikzstyle{block}=[draw, rectangle, rounded corners, minimum height=1.1cm, minimum width=2.8cm, align=center]

\node[block, draw=blue] (query) at (3,1.8) {Query: $W_Q$};
\node[block, draw=blue] (key) at (3,0) {Key: $W_K$};
\node[block, draw=green] (value) at (3,-1.8) {Value: $W_V$};
\node[block, draw=red] (score) at (7,1) {Attention Score};
\node[block, draw=black] (combo) at (7,-1) {Combined Value};

\node[] (X) at (0,0) {X};

\node[] (H) at (10,0) {H};
\draw[->,black] (combo.east) -|(9,0) -- (H);

\draw[->, green] (value.east) -| (5,-1.8) node[below, right] {V} |- (combo.west);
\draw[->, blue] (key.east) -| (5,0) node[below, right] {K} |- ([yshift=-0.2cm]score.west);
\draw[->, blue] (query.east) -| (5,1.8) node[above, right] {Q} |- ([yshift=0.2cm]score.west);
\draw[->, red] (score.south) -- (combo.north) node[midway, right] {Z};
\draw[->,black] (X.east) -- (key);
\draw[->,black] (1,0) |- (value);
\draw[->,black] (1,0) |- (query);





\end{tikzpicture}
}
\caption{
Information flow for Transformer Attention.
$X$ is the hidden state projected into query ($Q$), key ($K$), and value ($V$) matrices. Attention scores ($Z$) are used to weight $V$, producing the attended output ($H$).
}
\label{fig:transformer}
\end{figure}

Transformers rely on an attention mechanism to capture relationships between elements in a sequence and are constructed as a vertical stack that alternates attention and feed-forward blocks. Each attention block, in turn, contains multiple attention heads that process information, as illustrated in Figure \ref{fig:transformer}. As a first step, the hidden internal state $X$ is projected into the \textit{query}, \textit{key}, and \textit{value} matrices by a multiplication with their respective weight matrix, 
\begin{equation}
Q = X W_Q, \, K = X W_K, \, \mathrm{and} \, V = X W_V,     
\end{equation}
such that $Q,K,V \in \mathbb{R}^{L \times d}$ where  $L$ is the sequence length and $d$ is the latent dimension.

\paragraph*{Conventional Attention}
In the standard Transformer \cite{Vasvani2017}, attention scores are computed as the dot-product between the \textit{query} and \textit{key} matrices, followed by a Softmax normalization
\begin{equation}
Z_{ij} =  \underset{j}{\text{softmax}}\left(
    \frac{1}{\sqrt{d}} \sum_k Q_{ik} K_{jk}
    \label{eq:softmax_attention}
\right),
\end{equation}

These scores are then used to make a weighted combination of the \textit{value} matrix by multiplication
\begin{equation}
H_{ik} = \sum_j Z_{ij} V_{jk} 
    \label{eq:mult_combination}
\end{equation}

Alternative formulations such as Fastformer \cite{wu2021fastformer} and ReLUformer \cite{ReLUformer} introduce additive attention or replace Softmax with ReLU activation. While these approaches reduce the number of complex operations, they retain multiplication between attention and value matrices.

\paragraph*{Inhibitor Attention}
The literal multiplications that apply weights to the state $X$ to obtain the projects $Q$, $K$, and $V$ remain the same on the left side of Figure \ref{fig:transformer}. 

The variable-to-variable multiplication and Softmax normalization of eq.~\eqref{eq:softmax_attention} are replaced with  
\begin{equation}
    Z_{ij} = \frac{\gamma}{\sqrt{d}} \sum_k | Q_{ik} - K_{jk} |,
    \label{eq:inhibitor_attention}
\end{equation}
where $\gamma$ is a learnable scaling factor. This substitutes the dot-product similarity in conventional attention with a Manhattan distance-based formulation.

The attended output replacing eq.~\eqref{eq:mult_combination} is formed by subtractive inhibition rather than multiplicative weighting
\begin{equation}
    H_{ik} = \eta \sum_j \big( V_{jk}^+ - \tilde{Z}_{ij}^+ \big)^+ 
        + \eta \sum_j \big( V_{jk}^- + \tilde{Z}_{ij}^+ \big)^-,
    \label{eq:inhibitor_combination}
\end{equation}
where $(x)^+ = \max(x,0)$ and $(x)^- = \min(x,0)$ denote positive and negative ReLU components, and $\bar{Z}_{ij}$ is the centered and shifted inhibition score
\begin{equation}
    \tilde{Z}_{ij} =  Z_{ij} - \underset{j}{\text{mean}} \left( Z_{ij} \right) - \delta,
    \label{eq:inhibitor_centering}
\end{equation}
which compensates for the loss of normalisation provided by softmax in the conventional mechanism. The inhibitor design, therefore, introduces new learnable parameters, $\gamma$, $\eta$, and $\delta$, for each attention head. 

Feed-forward and normalization layers remain unchanged from the standard Transformer block.

\subsection{Gated Recurrent Neural Networks}

\begin{figure}
\centering
\resizebox{\columnwidth}{!}{%
    \begin{tikzpicture}[>=latex, thick]

\tikzstyle{block}=[draw, rectangle, rounded corners, minimum height=1.1cm, minimum width=2.8cm, align=center]

\node[block, draw=blue] (update) at (0,0) {Update Gate};
\node[block, draw=green] (proposed) at (0,-2.4) {Propose State};
\node[block, draw=red] (gating) at (-4,-1.2) {Gating Score};

\node[left=5.5cm of update] (htm1) {$h_{t-1}$};
\node[below=2.2cm of gating] (xt) {$x_t$};

\node[right=1cm of update] (ht) {$h_t$};

\draw[->, blue] (htm1) -- (update.west);
\draw[->, blue] (gating.west) ++(-0.5cm,1.2cm) -- ++(-0,-1.2cm) -- (gating.west);
\draw[->, blue] (update.west) ++(-0.5cm,0) -- ++(-0,-2.4cm) -- (proposed.west);

\draw[->] (xt) -- (gating.south);
\draw[->] (gating.south) ++(0,-1.75cm) -- ++(4cm,0) -- (proposed.south);

\draw[->, red] (gating.east) -- (-0.4cm,-1.2cm) -- ([xshift=-0.4cm]update.south) node[pos=0.25, left] {$z_t$};
\draw[->, green] ([xshift=0.4cm]proposed.north) -- ([xshift=0.4cm]update.south) node[midway, right] {$\hat{h}_t$};

\draw[->,blue] (update.east) -- (ht);

\end{tikzpicture}
}
\caption{
Information flow in a simplified gated RNN.
$x_t$ is the current input and $h_{t-1}$ the previous hidden state. The gating score ($z_t$) controls the combination of $h_{t-1}$ and the proposed update ($\hat{h}_t$), producing the new hidden state ($h_t$).
}
\label{fig:gated_rnn}
\end{figure}

Gated RNNs address the vanishing gradient problem by introducing a gating mechanisms that control how information is updated and retained over time. The information flow for processing a single step in a sequence is illustrated in Figure \ref{fig:gated_rnn}. 
Here, we consider a simplified gated RNN and write the update rule in terms of a \textit{gating score} 
\begin{equation}
    z_t = W_z x_t + U_z h_{t-1} + b_z,
    \label{eq:linear_score}
\end{equation}
where the parameters $W_z$, $U_z$, and $b_z$ are learned from the training data.
Note that for our exposition, the score is linear.

\paragraph*{Conventional Gate}
In standard gated RNN architectures such as LSTM~\cite{Hochreiter_1997} and GRU~\cite{Chung_2014}, the gating score $z_t$ is passed through a sigmoid activation to produce a gating term $\sigma(z_t) \in (0,1)$. This term determines the interpolation between the previous hidden state $h_{t-1}$, and a proposed candidate update $\hat{h}_t$. The update rule can be expressed as
\begin{equation}
    h_t = \sigma(z_t) \odot h_{t-1} + (1 - \sigma(z_t)) \odot \hat{h}_t,    
    \label{eq:GRU_update}
\end{equation}
where $\odot$ denotes element-wise multiplication. The proposed state $\hat{h}_t$ is typically computed as a nonlinear function $\phi$ of the current input and the previous state, $\hat{h}_t = \phi(x_t, h_{t-1})$, often involving additional gates such as a reset gate. 

These multiplicative interactions allow the model to regulate information flow and capture long-term dependencies effectively. However, they introduce computational overhead due to the sigmoid activation and variable-to-variable multiplications.

\paragraph*{Inhibitor Gate}
The linear score calculation of eq.~\eqref{eq:linear_score} remains the same, but we replace sigmoid activation with ReLU and elementwise multiplication by subtraction. 

The inhibitor update rule replacing eq.~\eqref{eq:GRU_update} combines the previous state $h_{t-1}$ and the proposed state $\hat{h}_t$ according to
\begin{equation}
    h_t = (h_{t-1} + z_t^-)^+ + (\hat{h}_t - z_t^+)^+,
    \label{eq:iGRU_update}
\end{equation}
where the negative and positive ReLU operations are denoted as defined for the inhibitor transformer. The inhibitor terms act as additive constraints, providing a mechanism for selective information retention and update.

The reset gate in GRU and the other gates for LSTM architectures are modified similarly, replacing multiplicative interactions with additive inhibition. Detailed formulations for all gates are omitted for brevity in this manuscript but can be found in the workshop paper \cite{poster2024_PPAI_gatedRNN}.

\section{Experiments}
\label{sec:experiments}

We evaluate the proposed architectures on standard benchmark tasks commonly used to assess sequence modeling and attention mechanisms. The experiments aim to test the functional training capacity of inhibitor mechanisms under practical constraints rather than optimize for state-of-the-art performance. 

Our setup includes two components: simple machine learning tasks—ranging from synthetic sequences to real-world vision and language benchmarks—and a knowledge distillation experiment designed to assess inhibitor-based Transformers on non-trivial natural language understanding tasks.

\subsection{Machine Learning Tasks}

We use four simple machine learning tasks that are widely employed to benchmark sequence models. Each experiment was repeated at least 20 times to estimate differences between the two mechanisms with statistical significance.

\paragraph*{Synthetic Adding Task}
Hochreiter \ introduced the adding problem \cite{Hochreiter_1997} to test the ability of models to capture long-term dependencies. Each input consists of two sequences of length 100: a sequence of random numbers and a two-hot sequence. The target output is the sum of two selected elements, equivalent to the dot product of the two sequences. This task is simple for models with effective memory mechanisms but challenging for architectures without gating or attention. For gated RNN experiments, sequences were processed using recurrent cells with either conventional or addition-based Inhibitor gates. For Transformer experiments, the same sequences were modeled using one-layer attention blocks with either conventional dot-product or inhibitor attention. 

The models were trained directly on the task to evaluate training capacity. The training set contains 20{,}000 examples, and the test set includes 5{,}000 examples. Note that the Adding task is also revisited in Section~\ref{sec:analysis} to assess scaling behavior under encrypted inference.

\paragraph*{MNIST}
The MNIST dataset \cite{lecun-mnisthandwrittendigit-2010} consists of handwritten digit images of size $28\times28$. For sequence-based processing, the image height is treated as a temporal dimension. This task evaluates the model’s ability to handle structured visual input. In gated RNN experiments, two recurrent layers were followed by a fully connected layer with batch renormalization. For Transformer experiments, a single attention layer was applied to the sequence representation of the image. Models were trained for 50 epochs.

\paragraph*{IAMW}
The IAM Words Handwriting Database \cite{IAM_database_2002} contains handwritten words from over 700 writers and is widely used for offline handwriting recognition. The task involves predicting character sequences from image input, requiring models to capture both spatial and sequential dependencies. For gated RNN experiments, the architecture combined convolutional layers with a two-level bidirectional recurrent network and used Connectionist Temporal Classification (CTC) loss \cite{Graves2006}. For Transformer experiments, convolutional layers were followed by an attention block. Edit distance \cite{Navarro2001} was used as the evaluation metric for this task. Models were trained for 50 epochs.

\paragraph*{IMDB}
The IMDB dataset \cite{imdb} consists of movie reviews labeled for sentiment polarity. This task evaluates the model’s ability to process text sequences and extract semantic information. For gated RNN experiments, bidirectional recurrent networks were trained over 10 epochs for sentiment classification. For Transformer experiments, we trained a one-layer model using either conventional dot-product attention or the proposed Inhibitor attention mechanism. The input text was tokenized and embedded into a fixed-length sequence representation, which was processed by a single attention block followed by a feed-forward network with ReLU activation and layer normalization. The evaluation metric was classification accuracy on the IMDB test set.

\subsection{Training Results for Transformers}

\begin{table} 
\centering
\small
\caption{
Comparison of training performance (on plaintext data) for conventional and inhibitor-based Transformers on four benchmark tasks: mse for Adding, accuracy for MNIST and IMDB, and edit distance for IAMW.
}
\begin{filecontents*}{results/all_results.txt}
Benchmark Tasks,Adding,MNIST,IMDB,IAMW
Conv. Transformer,0.0011,98.2\%,87.2\%,17.9
Inhib. Transformer,0.0012,97.9\%,87.3\%,18.1
\end{filecontents*}
\csvautobooktabular{results/all_results.txt}
\label{table:results_summary}
\end{table}

Test set performance for the simple sequence learning benchmark tasks is summarized in Table~\ref{table:results_summary}.
As shown in the table, the Inhibitor Transformer achieves performance very similar to the conventional Transformer across all tasks. Differences between the two mechanisms are small and not statistically significant at the 95\% confidence level.

Implementation details for the Transformer training experiments are available in~\cite{poster2024TransformerFHE}.

\subsection{Training Results for Gated RNNs}

\begin{table}[!t]
\caption{%
Comparison of training performance (on plaintext data) for conventional and inhibitor-based gated RNNs on four benchmark tasks: MSE for Adding, accuracy for MNIST and IMDB, and edit distance for IAMW.
}
\label{tab:rnn_comparison}
\centering
\begin{filecontents*}{results/rnn_results.txt}
Benchmark Tasks,Adding,MNIST,IMDB,IAMW
Conv. Gated RNN,0.005,99.3\%,86.9\%,17.4
Inhib. Gated RNN ,0.003,99.1\%,87.0\%,17.5
\end{filecontents*}
\csvautobooktabular{results/rnn_results.txt}
\end{table}

Table~\ref{tab:rnn_comparison} summarizes the training performance of conventional gated RNNs and the proposed addition-based inhibitor variants evaluated on the test sets for the four benchmark tasks.

For the {Synthetic Adding} task, the Adding column shows the mean squared error (MSE) over all trials. The inhibitor GRU performs comparably to the conventional GRU on this task. On {MNIST}, both architectures achieve high accuracy, with only minor differences between conventional and inhibitor designs.
For {IAM Words} handwriting recognition, inhibitor-based models perform slightly worse than conventional ones, with a statistically significant difference in edit distance.
Also on {IMDB} sentiment analysis, inhibitor-based RNNs show comparable accuracy to the conventional gated RNNs. For more details on the implementation, we refer to~\cite{poster2024_PPAI_gatedRNN}.

\subsection{Knowledge Distillation Setup}

\begin{table*} 
    \centering
    \caption{%
    Comparison of conventional DistilBERT and inhibitor-based DistilBERT pre-trained via task-agnostic knowledge distillation. Both models were fine-tuned on GLUE and IMDB benchmarks; results are averaged over three runs.
    }
    \begin{filecontents*}{results/distillation.txt}
Models,Score,CoLA,MNLI,MRPC,QNLI,QQP,RTE,SST-2,STS-B,WNLI,IMDB
Conv. DistilBERT,77.0,51.3,82.2,87.5,89.2,88.5,59.9,91.3,86.9,56.3,92.82
Inhib. DistilBERT,74.5,40.0,79.2,86.8,85.4,89.5,59.2,90.2,83.5,56.3,92.81
\end{filecontents*}
\csvautobooktabular{results/distillation.txt}
    \label{tab:model-comparison}
\end{table*}

Knowledge distillation (KD) \cite{Hinton2015} is a model compression technique that transfers knowledge from a larger teacher model to a smaller student model, enabling reduced computational cost without substantial accuracy loss. 
Our experiments were based on the DistilBERT \cite{DistilBERT} paper, but instead of using a full-sized BERT model as the teacher, we used the smaller pre-trained DistilBERT model for computational convenience and simpler alignment.
The weights were also initialized from the teacher model.
We only discuss task-agnostic KD here.  We refer to~\cite{poster2025inhibidistilbert} for detailed information on the experiments, including hyperparameter listings as well as supplementary results for task-specific KD. 

The initial phase involved layerwise training to align the contextual representations between the teacher and student models using 10\% of the Wikitext-103 corpus. In this phase, all weights in the student model were frozen except for the weight matrices of the query, key, and value components in the current layer being trained. The Mean Squared Error (MSE) loss function was applied to align the context outputs of corresponding layers in both models. Each layer was trained iteratively from the bottom to the top layer (layer 0 to layer 5). After training a layer for two epochs, its weights were frozen, and the next layer in the sequence was unfrozen.

Following the layerwise training, a full-layer training phase was conducted using 60\% of the Wikipedia 20231101 corpus. In this phase, all layers in the student model were unfrozen, allowing parameter updates across the entire network. MSE loss was applied to the hidden states to align the hidden layer outputs between the teacher and student models.

Once the task-agnostic knowledge distillation was completed, the final weights of the inhibitor DistilBERT model were stored and used as the foundation for the fine-tuning tasks. 

\paragraph*{GLUE Finetuning}
We evaluate our inhibitor-based DistilBERT model against the conventional scaled dot-product attention DistilBERT on the GLUE benchmark \cite{wang2019gluemultitaskbenchmarkanalysis}, which consists of nine language understanding tasks such as natural language inference (MNLI), question answering (QNLI), paraphrase detection (QQP), and sentiment classification (SST-2). Each model was fine-tuned for three epochs using the AdamW optimizer with standard hyperparameters, following the practices outlined in the original DistilBERT paper \cite{DistilBERT}. The goal of this experiment is to determine whether the inhibitor attention mechanism can maintain competitive performance when integrated into a compressed Transformer architecture.

\paragraph*{IMDB Finetuning}
We also fine-tuned both models on the IMDB sentiment analysis dataset. This task provides a focused evaluation of the inhibitor mechanism on a single-domain text classification problem. Fine-tuning was performed using the same optimization strategy as for GLUE, and results are reported as classification accuracy on the IMDB test set.

\subsection{Training Results for Knowledge Distillation}

We evaluate our inhibitor-based DistilBERT model and the conventional scaled dot-product attention DistilBERT, both finetuned against the nine different language understanding tasks of the GLUE benchmark \cite{wang2019gluemultitaskbenchmarkanalysis}. 
For reference, results for the IMDB sentiment analysis task are also presented. 

The performance comparison in Table \ref{tab:model-comparison} indicates that a fine-tuned inhibitor DistilBERT achieves competitive accuracy, with a modest 3.2\% average drop on GLUE compared to dot-product DistilBERT across the different tasks. 
All reported scores are averaged over three runs to reduce variance.
We note a larger performance drop for Inhibitor DistilBERT on the CoLA task.

\section{Scaling Analysis}
\label{sec:analysis}

While the training experiments in the previous section indicate that inhibitor-based architectures achieve functional equivalence to conventional designs, efficiency across different execution settings remains important. This section examines computational scaling for both plaintext and encrypted inference and analyzes how the architectural changes affect complexity, circuit depth, and execution time.

\subsection{Computational Complexity}

\begin{table*}
\centering
\caption{%
Asymptotic complexity of key operations for one RNN update or one Transformer attention head for conventional and inhibitor mechanisms. Here, $n$ and $m$ denote RNN state and input size; $L$ and $d$ denote Transformer sequence length and hidden dimension.
}
\label{tab:complexity_comparison}
\begin{tabular}{lcccc}
\toprule
\textbf{Operation Type} & \textbf{Conv Gated RNN} & \textbf{Inhb Gated RNN} & \textbf{Conv Attention} & \textbf{Inhb Attention} \\
\midrule
Literal multiplication (weights) & $\mathcal{O}(n^2+n\cdot m)$ & $\mathcal{O}(n^2+n\cdot m)$ & $\mathcal{O}(L\cdot d^2)$ & $\mathcal{O}(L\cdot d^2)$ \\
Variable additions or subtractions & $\mathcal{O}(n)$ & $\mathcal{O}(n)$ & $\mathcal{O}(L^2\cdot d)$ & $\mathcal{O}(L^2\cdot d)$ \\
Univariate function evaluations & $\mathcal{O}(n)$ & $\mathcal{O}(n)$ & $\mathcal{O}(L^2)$ & $\mathcal{O}(L^2)$ \\
Variable-to-variable multiplication & $\mathcal{O}(n)$ & $0$ & $\mathcal{O}(L^2 \cdot d)$ & $0$ \\
Variable-to-variable division & $\mathcal{O}(n)$ & $0$ & $\mathcal{O}(L^2)$ & $0$ \\
\bottomrule
\end{tabular}
\end{table*}

Table~\ref{tab:complexity_comparison} summarizes the asymptotic complexity of key operations for one update in gated RNNs and one attention head in Transformers, comparing conventional and inhibitor-based mechanisms. Here, $n$ denotes the RNN hidden state size, $m$ the input size, $L$ the Transformer sequence length, and $d$ the hidden dimension.

The rows are ordered by increasing computational cost under TFHE, where operations using bootstrapping dominate.
Specifically, univariate function evaluations require a single PBS, variable-to-variable multiplications require two PBS operations, and divisions typically involve three PBS operations, making them the most expensive.

\begin{description}
    \item[Literal multiplication (weights):] Multiplications between plaintext weights and encrypted variables. For gated RNNs, this includes input-to-hidden and hidden-to-hidden projections, giving $\mathcal{O}(n^2 + n \cdot m)$. For Transformers, producing Q, K, and V requires $\mathcal{O}(L \cdot d^2)$ operations. These multiplications are relatively inexpensive under TFHE because one operand is plaintext.

    \item[Variable additions or subtractions:] Combining states in RNNs or applying attention weights to values in Transformers. For RNNs, this is $\mathcal{O}(n)$ per update, while for Transformers it scales as $\mathcal{O}(L^2 \cdot d)$ because each of the $L$ queries aggregates over $L$ keys for $d$ dimensions.

    \item[Univariate function evaluations:] This includes ReLU as well as the use of the exponential function for the evaluation of the sigmoid/softmax activation functions. Under TFHE, these are implemented via programmable bootstrapping. Complexity is $\mathcal{O}(n)$ for RNNs and $\mathcal{O}(L^2)$ for Transformers of both varieties.

    \item[Variable-to-variable multiplication:] Present in conventional designs for gating (RNN) and dot-product attention (Transformer). Complexity is $\mathcal{O}(n)$ for RNNs and $\mathcal{O}(L^2 \cdot d)$ for Transformers. Removed for the inhibitor designs.

    \item[Variable-to-variable division:] Occurs in normalization steps such as softmax and sigmoid. Complexity is $\mathcal{O}(n)$ for RNNs and $\mathcal{O}(L^2)$ for Transformers. Removed in inhibitor-based designs.
\end{description}

Additionally, for the Transformer, the feed-forward layer applies two linear projections and an activation function to each of the $L \times d$ inputs. These projections involve only literal multiplications and additions, with complexity $\mathcal{O}(L \cdot d^2)$. The activation step requires one PBS per element, adding $\mathcal{O}(L \cdot d)$ to the cost. Layer normalization has complexity $\mathcal{O}(L \cdot d)$ and is dominated by attention in overall cost, but its reliance on variable-to-variable multiplication and division requires repeated applications of PBS under TFHE.

In summary, while the dominant asymptotic term for both architectures remains unchanged—$\mathcal{O}(n^2 + n \cdot m)$ for gated RNNs and $\mathcal{O}(L \cdot d^2 + L^2 \cdot d)$ for Transformer attention — the inhibitor approach eliminates the operations that are disproportionately expensive under TFHE.

\subsection{TFHE Implementation Details}
For all cipher text experiments, we use the Concrete Python library for TFHE~\cite{Concrete_Python_Compiler}. 
To ensure correct circuit evaluation, all component values must fit within the allotted message space. TFHE operates on several ciphertext types—LWE \cite{Regev2009}, RLWE \cite{Lyubashevsky2013_RLWE}, GLWE \cite{Brakerski_2012}, and GSW generalizations \cite{Gentry2013_GSW}—each with specific parameters. 
Parameter optimization frameworks~\cite{Bergerat2023_ParamOpt} distinguish macro-parameters (ciphertext structure) from micro-parameters (decomposition base, PBS levels), relying on noise and cost models to ensure correctness and minimize overhead. In this work, all parameters are automatically selected using the compiler.

As the Concrete library only supports integers natively, quantization becomes necessary when implementing neural network architectures: inputs, weights, and activations all need to be projected from floating-point values onto the integers. Furthermore, since any deep neural network with non-trivial activation functions requires PBS, the maximum precision of the TFHE table look-up implementation will be a limiting factor. At the time of the experiments, this was $2^7=128$ different values, i.e.,\ integer 7-bit precision. 
Precision also directly impacts execution time as larger table lookups imply slower inference.

\subsection{Scaling Experiments for Transformers}

\begin{table} 
\centering
\small
\caption{%
Estimated plaintext execution time on CPU for conventional and inhibitor-based Transformers across four sequence lengths (single attention head).
}
\begin{filecontents*}{results/timing_short.txt}
Timing Plaintext,32,64,128,256
Conv. Transformer,98.6 µs,330 µs,1.2 ms,4.48 ms
Inhib. Transformer,63.1 µs,178 µs,577 µs,2.5 ms
\end{filecontents*}
\csvautobooktabular{results/timing_short.txt}
\label{table:results_timing}
\end{table}

\begin{table} 
\centering
\small
\caption{%
Estimated encrypted execution time on CPU for conventional and inhibitor-based Transformers across four sequence lengths (single attention head).
}
\begin{filecontents*}{results/timing_short_beta4_2_0_2_2.txt}
Timing Encrypted,2,4,8,16
Conv. Transformer,2.68 s,22.4 s,107 s,828 s
Inhib. Transformer,0.749 s,8.56 s,23.8 s,127 s
\end{filecontents*}
\csvautobooktabular{results/timing_short_beta4_2_0_2_2.txt}
\label{table:results_encrypted}
\end{table}

We compare execution time scaling for inhibitor-based attention and conventional dot-product attention under plaintext and encrypted settings. For plaintext experiments, attention mechanisms were implemented in low-level code using 16-bit integer arithmetic, and timing was measured on CPU with the Criterion benchmarking framework. For encrypted experiments, we used the TFHE scheme via the Concrete library, compiling circuits for integer operations with up to 8-bit precision. Quantization was necessary to fit within TFHE lookup table limits.

Execution time increases with sequence length for both mechanisms, but inhibitor attention consistently outperforms conventional attention. In plaintext settings, inhibitor achieves a 30–50\% reduction in execution time across all tested lengths. Under TFHE encryption, the improvement is even more pronounced, with speedups ranging from 3 to 6 times compared to conventional dot-product attention.

Further implementation details can be found in~\cite{poster2024TransformerFHE}.

\subsection{Scaling Experiments for Gated RNNs}

\begin{table} 
\centering
\small
\caption{
Mean encrypted execution time on CPU (100 trials) for gated RNNs solving the Adding and Copying Memory tasks. Comparison of additive inhibitor gate ($\mathrm{Add}$) and conventional multiplicative gate ($\mathrm{Mul}$) for different quantization precision.
}
\begin{filecontents*}{results/timing_task1and2.txt}
,Add 4b,Mul 1b,Mul 2b,Mul 3b,Mul 4b
Adding,0.50 s,0.44 s,0.62 s,2.03 s,31.5 s
Copying,1.13 s,1.61 s,2.33 s,10.1 s,104 s
\end{filecontents*}
\csvautobooktabular{results/timing_task1and2.txt}
\label{table:timing_task1and2}
\end{table}

\begin{table} 
\centering
\small
\caption{
Mean encrypted execution time on CPU (100 trials) for one step of a gated RNN with 
$n$-bit precision for input and state. Comparison of additive inhibitor gate ($\mathrm{Add}$) and conventional multiplicative gate ($\mathrm{Mul}$) for different quantization precision.
}
\begin{filecontents*}{results/timing_task3_m1.txt}
n,Add 4b,Mul 1b,Mul 2b,Mul 3b,Mul 4b
2,0.13 s,0.13 s,0.21 s,0.26 s,0.63 s
3,0.18 s,0.19 s,0.26 s,0.60 s,2.02 s
4,0.22 s,0.22 s,0.62 s,2.10 s,5.57 s
5,0.52 s,0.53 s,2.03 s,5.56 s,25.6 s
6,1.76 s,1.79 s,5.48 s,25.9 s,
7,4.64 s,4.84 s,26.4 s,,
8,21.1 s,22.5 s,,,
\end{filecontents*}
\csvautobooktabular{results/timing_task3_m1.txt}
\label{table:timing_task3}
\end{table}

We evaluate the scaling behavior of gated RNNs under encrypted execution. \textbf{Note:} Plaintext scaling experiments are not reported here because weight multiplications dominate execution time for both conventional and inhibitor designs (see Table~\ref{tab:complexity_comparison}), resulting in no meaningful difference. Refer to~\cite{poster2023gatedRNN_unpublihed, poster2024_PPAI_gatedRNN} for implementation details.

We use two synthetic sequence tasks: the Adding problem (described in Section~\ref{sec:experiments}) and the Copying Memory problem introduced by Hochreiter~\cite{Hochreiter_1997}. The latter is a standard benchmark for assessing long-term memory in recurrent architectures. It consists of a sequence of $k$ random symbols drawn from $n$ distinct tokens, followed by $T-1$ blanks and recall markers. After observing the recall signal, the model must reproduce the initial sequence, requiring retention over long delays. In our experiments, both tasks were implemented for gated RNNs using handcrafted weights, with linear activation for gating and ReLU for the update proposal.

Two scenarios were evaluated. First, execution time under TFHE was measured for one step of the gated RNN on the Adding and Copying Memory tasks using different quantization settings for the conventional gate, while the addition-based gate operates without quantization loss. Results, averaged over 100 trials, are shown in Table~\ref{table:timing_task1and2}. The column labels also indicate the precision used for sigmoid quantization in the multiplicative gate case. All experiments use handcrafted weights.

Second, we benchmarked a simplified scalar gated RNN with 1-bit weights and varying $n$-bit precision for input and state. Clip functions were applied to constrain values within an allowed range, enabling combination according to either the addition-based or multiplication-based rule. Experiments were repeated for randomly sampled inputs, states, and weights, as well as for all combinations of minimum and maximum values. Table~\ref{table:timing_task3} reports mean execution times for one step under these conditions, averaged over 100 trials. The first column presents results for the additive inhibitor architecture, while the remaining columns show the conventional multiplicative gate at different quantization precision.

Across both experiments, execution time for the conventional gate grows rapidly with increasing precision, whereas the addition-based inhibitor gate maintains significantly lower execution times across all tested configurations.

\section{Discussion}
\label{sec:discussion}

This section discusses the results in relation to our central question: whether inhibitor-based architectures that eliminate multiplications and complex activations can enable practical encrypted inference without compromising accuracy.

\subsection{Key Findings}
The empirical evaluation demonstrates that inhibitor-based architectures achieve accuracy comparable to conventional mechanisms across both elementary benchmarks (Adding, MNIST, IAM handwriting, IMDB sentiment analysis) and more complex natural language understanding tasks represented by GLUE. These experiments were designed to assess functional equivalence under standard training conditions rather than to achieve state-of-the-art performance. Statistical analysis indicates that observed differences between inhibitor and conventional designs are not significant at the 95\% confidence level.

In terms of computational efficiency, the proposed architectures exhibit substantial improvements. Specifically, inhibitor-based models deliver a 30--50\% reduction in execution time under plaintext conditions and a 3--6 times acceleration under TFHE encryption. 

We next analyze the design principles that enable these improvements.

\subsection{Design Insights}
We note that these reductions in inference time are achieved without any change in model size -- this distinguishes the inhibitor approach from model pruning techniques \cite{Han2015, Han2016} that reduce execution time by removing weights and thereby decreasing the number of operations.

The efficiency improvements can be attributed to the systematic elimination of operations that dominate computational cost under homomorphic encryption. Inhibitor designs replace variable-to-variable multiplications and complex activation functions (Softmax, Sigmoid) with addition and ReLU, thereby reducing circuit depth and bootstrapping overhead. The complexity analysis in Table~\ref{tab:complexity_comparison} confirms that these modifications remove the most expensive operations under TFHE—namely multiplications and divisions—which require multiple programmable bootstrapping steps.

The inhibitor Transformer replaces scaled dot-product attention (related to cosine similarity) with Manhattan distance. This can be interpreted as a change of the metric used to quantify distance/similarity between the embedded token representations.

We also note that the inhibitor design introduces new learnable parameters $\gamma$, $\eta$, and $\delta$ that compensate for the loss of the normalisation provided by softmax. This is only a modest expansion of the model size with three parameters per attention head and is completely negligible compared to the memory footprint of the weights for the embedding, attention, and feedforward blocks. 

Furthermore, inhibitor attention and gating mechanisms support end-to-end integer-only computation, greatly simplifying encrypted inference under TFHE. This represents an extreme form of neural network quantization \cite{Zhou2016,Jacob_2018,Krishnamoorthi2018}, which also in plaintext settings reduces memory footprint and accelerates execution by employing low-bit fixed-point arithmetic instead of floating-point operations.

Nevertheless, certain components, such as layer normalization, continue to rely on multiplication and division. These operations remain potential candidates for inhibitor-style redesign, offering an avenue for further optimization in encrypted environments.

Having established the rationale, we position our approach relative to prior work.

\subsection{Comparative Positioning}
Relative to alternative approaches such as Fastformer \cite{wu2021fastformer} and ReLUformer \cite{ReLUformer}, the inhibitor transformer provides a more radical simplification by eliminating all variable-to-variable multiplications from the attention mechanism. This characteristic positions it as particularly suitable for privacy-preserving computation under TFHE. 

Early approaches such as Cryptonets \cite{CryptoNets2016} relied on leveled homomorphic encryption and polynomial activations, which severely limited scalability. More recent TFHE-based methods modeled vision tasks with conventional neural network structures—convolutions \cite{Stoian2023} and fully connected layers \cite{Selvakumar2025}—and adapted them for TFHE constraints using quantization-aware training and programmable bootstrapping (PBS). This enables arbitrary depth networks with pruning strategies to control accumulator size. We note that CNN and FCNN mostly involve literal multiplications between encrypted values and plaintext weights. Our work is complementary and focuses on architectural redesign for sequence models, introducing inhibitor-based mechanisms for Transformers and gated RNNs that eliminate variable-to-variable multiplications.

Beyond algorithmic redesign, system-level optimizations offer complementary benefits. For example, GPU-accelerated TFHE bootstrapping achieves up to 20 times speedup \cite{Xiao2025}, and combining inhibitor-based designs with such hardware acceleration would compound efficiency gains, making encrypted inference more practical for real-world deployments..

Although the inhibitor design was motivated by the computational challenges of homomorphic encryption, replacing multiplications and complex activations with addition and ReLU also makes it suitable for resource-constrained deployments. These changes reduce arithmetic and memory overhead even in plaintext settings, lowering energy consumption on edge devices \cite{Horowitz_2014}, particularly on platforms without ML-specific accelerators.

Combining architectural co-design with quantization \cite{Zhou2016, Jacob_2018, Krishnamoorthi2018} and knowledge distillation \cite{Hinton2015, DistilBERT}, as explored in this work, aligns with widely adopted strategies for efficient model deployment, including TinyML and conditional computing.
This dual-purpose design enables privacy-preserving computation under encryption while providing a lightweight alternative also for plaintext deployments where power, latency, and memory are critical constraints.

\subsection{Limitations}

The study also reveals certain limitations. A notable drop in performance on the CoLA task in the GLUE benchmark suggests that inhibitor-based attention may struggle to capture syntactic nuances. Furthermore, the knowledge distillation pipeline employed a smaller teacher model (DistilBERT), which may account for the modest average performance gap on GLUE and perhaps also explain the CoLA underperformance. Future work should explore distillation from full-size BERT models and investigate improved strategies for transferring syntactic knowledge.

Direct transfer learning experiments from a conventional Transformer (BERT) to inhibitor-based architectures of equivalent size were unsuccessful, primarily due to incompatibilities in the attention mechanism that preclude weight reuse. This limitation motivated adopting knowledge distillation as a more effective strategy for transferring contextual representations. Although our long-term objective is to pre-train a full-sized inhibitor-based language model, resource constraints within the current study precluded this effort.

Hardware considerations must also be acknowledged: plaintext performance gains are modest on commodity server-grade CPU and quickly diminish on GPUs optimized for floating-point matrix multiplication (results not reported), underscoring the need for hardware co-design tailored to addition/ReLU-centric architectures.

\subsection{Future Directions}
The proposed inhibitor mechanisms offer potential advantages for deployment in privacy-preserving and resource-constrained environments. The limitations discussed above, however, motivate several future research directions:

\begin{description}
    \item[Hardware co-design:] Develop architectures optimized for addition and ReLU operations to minimize latency and energy consumption also for plaintext.
    \item[Normalization redesign:] Eliminate multiplication and division in normalization layers to enhance TFHE efficiency.
    \item[Pre-training:] Train a full-sized quantized inhibitor-based language model for large-scale benchmarking.
    \item[Knowledge distillation:] Refine distillation strategies to improve syntactic and semantic transfer across tasks.
    \item[Modern recurrence integration:] Extend inhibitor mechanisms to architectures such as xLSTM and HRM.
    \item[Broader evaluations:] Validate generality through multilingual datasets, vision applications, and generative models.
\end{description}

Hardware co-design and normalization redesign are immediate priorities to address computational bottlenecks, while full-scale pre-training and broader evaluations will help validate scalability. These efforts are driven by the need for privacy-preserving AI that enables ethical and compliant deployment in sensitive domains such as healthcare, finance, and public services, reducing risks of data misuse.

\section{Conclusion}
\label{sec:conclusion}

This work introduces inhibitor-based architectures as a practical alternative to conventional Transformers and gated RNNs for privacy-preserving machine learning under TFHE. By replacing costly multiplications and complex activations with addition and ReLU, these designs maintain competitive accuracy across diverse benchmarks while achieving efficiency gains of up to 50\% in plaintext execution and up to 6 times under TFHE encryption—without increasing model size. Their support for integer-only computation makes them promising architecture choices for encrypted inference and resource-constrained environments. 
Future work should address hardware co-design, improved distillation strategies, and scaling to full-sized language models to assess the practical potential of inhibitor-based architectures further.

\section*{Acknowledgment}
This research was supported by Vinnova under competitive research grants. Preliminary results were presented at AAAI-24, the 5th AAAI Workshop on Privacy-Preserving Artificial Intelligence (PPAI-24), the 3rd Annual FHE.org Conference, and the ICLR Workshop on Sparsity in LLMs (SLLM). The authors gratefully acknowledge constructive feedback and insights provided by participants and reviewers at these venues.

\bibliographystyle{IEEEtran}
\bibliography{main}

\end{document}